\documentclass[sigconf,authorversion,nonacm]{acmart}

\usepackage{multirow}
\usepackage{balance}
\AtBeginDocument{%
  }

\newcommand{\etal}{\textit{et al}. }
\newcommand{\ie}{\textit{i}.\textit{e}., }

\begin{document}

\title{ELMformer: Efficient Raw Image Restoration with a Locally Multiplicative Transformer}


\author{Jiaqi Ma}
\authornote{This work was done during Jiaqi Ma's internship at Horizon Robotics.}
\affiliation{%
  \institution{School of Computer Science, Wuhan University}
  \country{Wuhan, China}
}
\email{jiaqima@whu.edu.cn}

\author{Shengyuan Yan}
\affiliation{%
  \institution{School of Computer Science, Wuhan University}
  \country{Wuhan, China}
}
\email{shengyuan_yan@whu.edu.cn}

\author{Lefei Zhang}
\authornote{Corresponding author. He is affiliated with National Engineering Research Center for Multimedia Software, School of Computer Science, Wuhan University}
\affiliation{%
  \institution{Wuhan University}
  \institution{Hubei Luojia Laboratory}
  \country{Wuhan, China}
}
\email{zhanglefei@whu.edu.cn}

\author{Guoli Wang}
\affiliation{%
	\institution{Horizon Robotics}
	\country{Beijing, China}
}
\email{guoli.wang@horizon.ai}

\author{Qian Zhang}
\affiliation{%
	\institution{Horizon Robotics}
	\country{Beijing, China}
}
\email{qian01.zhang@horizon.ai}


\begin{abstract}
	In order to get raw images of high quality for downstream Image Signal Process (ISP), in this paper we present an Efficient Locally Multiplicative Transformer called ELMformer for raw image restoration. ELMformer contains two core designs especially for raw images whose primitive attribute is single-channel. The first design is a Bi-directional Fusion Projection (BFP) module, where we consider both the color characteristics of raw images and spatial structure of single-channel. The second one is that we propose a Locally Multiplicative Self-Attention (L-MSA) scheme to effectively deliver information from the local space to relevant parts. ELMformer can efficiently reduce the computational consumption and perform well on raw image restoration tasks. Enhanced by these two core designs, ELMformer achieves the highest performance and keeps the lowest FLOPs on raw denoising and raw deblurring benchmarks compared with state-of-the-arts. Extensive experiments demonstrate the superiority and generalization ability of ELMformer. On SIDD benchmark, our method has even better denoising performance than ISP-based methods which need huge amount of additional sRGB training images. The codes are release at \href{https://github.com/leonmakise/ELMformer}{https://github.com/leonmakise/ELMformer}.
\end{abstract}


\maketitle

\begin{figure}[ht]
	\centering
	\includegraphics[width=1\linewidth] {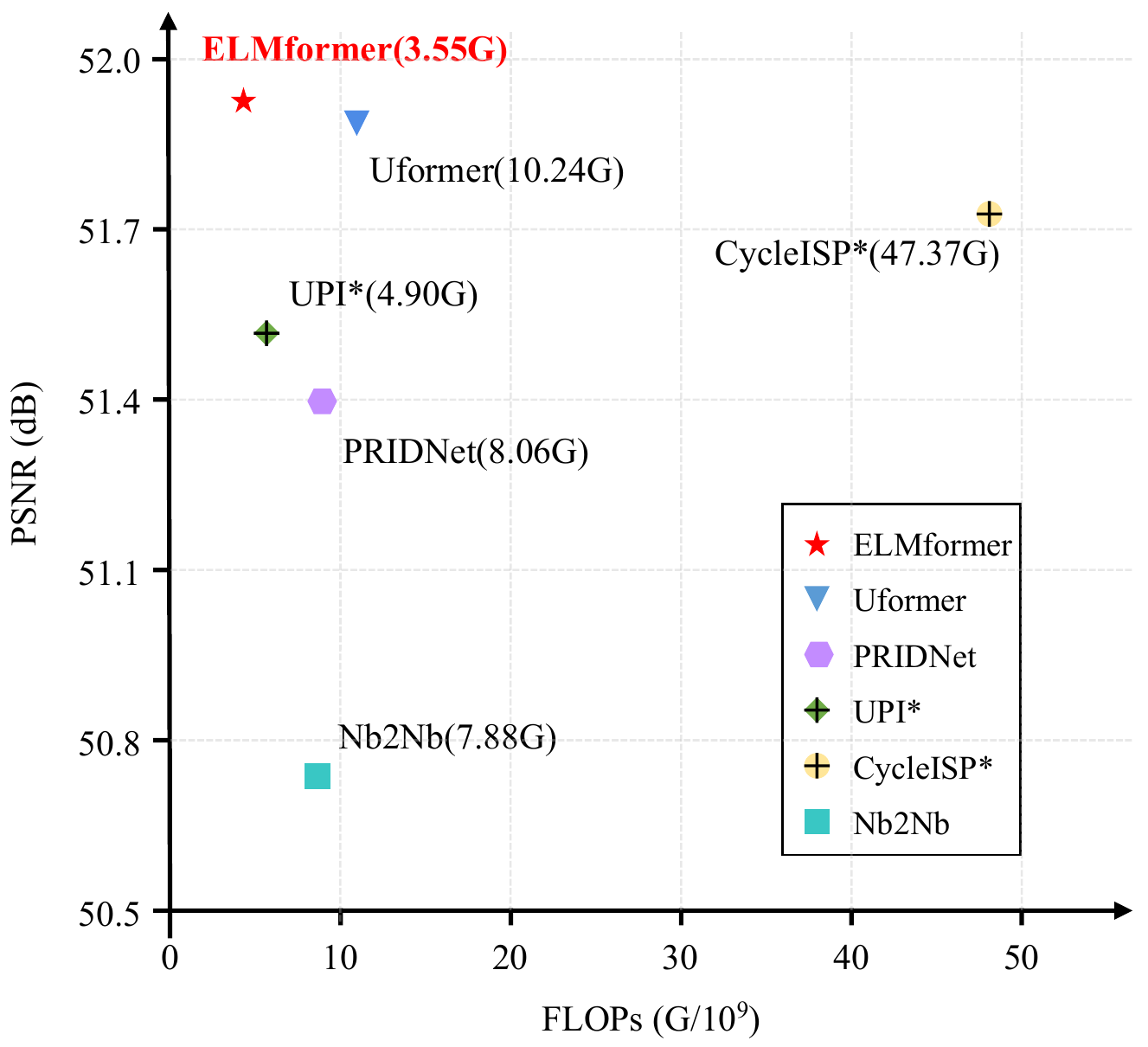}
	\caption{FLOPs and PSNR of denoising methods on SIDD benchmark with $128 \times 128$ patches. * represents that UPI and CycleISP use additional huge sRGB images for training, the others are trained on the SIDD Medium dataset.}
	\label{fig:top}
\end{figure}

\section{Introduction}
\label{sec:intro}
Image restoration is a classic low-level vision task in which raw image restoration is a special but important subtask. From the perspective of Image Signal Processing (ISP), raw data refers to unprocessed information from image sensors and is mainly captured as the initial data source. Raw restoration models output high-quality raw data whose noises and artifacts are removed. Restoration on raw images can be very essential because those final noisy sRGB images are processed by ISP which contains nonlinear transformation that will extort original shot-and-read noise. Non-negligible shot-and-read noises through the ISP pipeline result in the complex noise distribution of raw images, which can significantly obstacle restoration process. Hence, researchers turn to RAW data where noise and artifacts are uncorrelated and less complex \cite{DBLP:conf/dphoto/GhimpeteanuBSB16,DBLP:conf/cvpr/ZamirAKHK0020}.

Although recent models discuss the data structure and noise distribution of raw images, they still neglect the local context information hidden in the raw data structure. Hence, the raw image restoration task faces two main challenges. The first is insufficient exploitation of the raw images by the simple 'packing' strategy. The second is the balance between performance and computational cost caused by high-resolution raw images.

For the first challenge, researchers focus on pre-processing strategies to avoid damaging structures of raw images. Chen \etal \cite{DBLP:conf/cvpr/ChenCXK18} firstly connect raw images with sRGB ones by CNNs, and introduce a 'packing' strategy. Through 'packing', the single-channel raw image can be divided into four channels as 'RGGB' (Red, Green, Green, Blue) in Bayer patterns. Each channel is half-sized and only contains single color information. Hence, the packed raw images are color-independent in every channel and still keep the spatial dependencies around pixels' neighborhoods. Depending on this simple strategy, many works \cite{DBLP:conf/eccv/LiangGGZZ20, DBLP:conf/cvpr/XingE21, DBLP:journals/tip/LuJ22} related to raw images are proposed. However, Liu \etal \cite{DBLP:conf/cvpr/LiuWWXZHWCDFW19} reveal the error-prone mode when adapting augmentation methods designed for sRGB images to raw images. The errors easily occur especially when converting raw images to sRGB color space with cropping and flipping. Nevertheless, we notice that raw images are pixel-wise aligned in raw restoration tasks. Simple cropping or flipping operations would not harm the Bayer patterns, and the colors of restored sRGB images are consistent with degraded ones after demosaicing. Although the packing strategy has been a common sense in raw image processing, the spatial dependencies are actually weak. Liang \etal \cite{DBLP:journals/tmm/LiangCLH22} utilize Bi-directional Cross-modal Attention in every stage to enhance both color and spatial branches. From our insights, we prefer to leverage the advantages of color structures and spatial dependencies simultaneously for Transformer-based architectures, and we should ensure single-channel raw images to larger channels by projecting them into multiple subspaces. Hence, our ELMformer replace simple projection module with the Bi-directional Fusion Projection (BFP) module at the input stage.

In terms of the second challenge, we try to make use of those limited data by efficient and effective architectures. Zheng \etal \cite{DBLP:conf/cvpr/ZhengY021} propose a deep convolutional dictionary learning method to learn priors for both representation coefficients and dictionaries. Hu \etal \cite{DBLP:conf/cvpr/HuMLCZZW21} build a pseudo 3-D auto-correlation attention block through 1-D convolutions and a light-weight 2-D structure. Chang \etal \cite{DBLP:conf/eccv/ChangLFX20} design a residual spatial-adaptive block for denoising. Most focus on the noise modeling rather than the basic structural characteristics of raw data. With Transformer-based methods uprising, those methods outperform CNN-based methods and show their promising future. Wang \etal \cite{DBLP:journals/corr/abs-2106-03106} propose an U-shaped Transformer-based network for low-level tasks. Liang \etal \cite{DBLP:journals/corr/abs-2108-10257} device a residual architecture by Swin Transformer. Besides, some studies refer to generating adequate raw data pairs from existing sRGB images. They simulate the ISP pipeline into an invertible procedure to convert sRGB images into raw ones, \ie UPI \cite{DBLP:conf/cvpr/BrooksMXCSB19}, CycleISP \cite{DBLP:conf/cvpr/ZamirAKHK0020} and PseudoISP \cite{DBLP:journals/corr/abs-2103-10234}. However, the balance of performance and computational cost caused by high-resolution raw images is still a problem, especially for Transformer-based methods like Uformer \cite{DBLP:journals/corr/abs-2106-03106} and SwinIR \cite{DBLP:journals/corr/abs-2108-10257}. They only consider the utilization of Transformer blocks (designed for sRGB images) but ignoring the characteristics of raw data structure. Besides, the limited size of shifted windows results in small receptive fields. Simply enlarging window size will increase the computational consumption quadratically. Therefore, an effective strategy should ensure a larger receptive field and acceptable computational cost. As is shown in Fig. \ref{fig:top}, with the proposed Lm-Win Transformer blocks, our ELMformer balance the performance and FLOPs well compared with other SOTAs.

To conclude, we utilize the Bi-directional Fusion Projection (BFP) module and Locally multiplicative Window (Lm-Win) Transformer blocks especially for raw image restoration. Both BFP module and Lm-Win Transformer block are suitable for raw data's characteristics and can guarantee their efficiency and effectiveness simultaneously. Overall, we summarize the contributions of this paper as follows:

\begin{itemize}
	\item We analysis the weaknesses of 'packing' strategy on raw data and utilize a Bi-directional Fusion Projection module to generate the initial projected features. The BFP module fuses color and spatial information and provides efficient structure priors inside Bayer pattern. Besides, the BFP module enlarges the receptive field and reduces the computation cost at the same image patch scale.
	\item To further construct the dependencies between pixels, we propose an effective Locally Multiplicative SA (L-MSA) module to enhance the neighborhood dependencies in local sub-windows. The L-MSA module gives a specialized consideration of the characteristics of raw image data and is embedded in every Lm-Win Transformer block.
	\item Both quantitative and qualitative results demonstrate the superiority of our proposed method. Ablation studies prove the effectiveness of each proposed module in our method. The generalization ability is also proved in this paper.
\end{itemize}

\section{Related work}
\subsection{Image Restoration Architecture}
Image restoration is often referred to images in sRGB color space, while raw images are commonly excluded. When concerning denoising which is a principal subtask in image restoration, \cite{DBLP:conf/cvpr/0002HWZ21,DBLP:journals/tip/ZhangZCM017,DBLP:conf/cvpr/HuMLCZZW21,DBLP:conf/icml/LehtinenMHLKAA18} corrupt clean sRGB images with AWGN and other synthetic noise to supervise the training stage. \cite{DBLP:conf/cvpr/KrullBJ19,DBLP:conf/cvpr/ChengWHLFL21,DBLP:journals/access/Gurrola-RamosDA21} employ U-net architecture to conduct image denoising. Others \cite{DBLP:conf/cvpr/HuangLJLL21,DBLP:conf/nips/LaineKLA19,DBLP:conf/eccv/Wu0CRZ20} are self-supervised models. And for deblurring which is also essential for restoration, Nah \etal \cite{DBLP:conf/cvpr/NahKL17} propose a network which uses multi-scale images as input. GAN and RNN are also employed in \cite{DBLP:conf/cvpr/KupynBMMM18,DBLP:conf/cvpr/TaoGSWJ18}. Lu \etal \cite{DBLP:conf/cvpr/LuCC19} device disentangled representation for self-supervised deblurring. Liang \etal \cite{DBLP:conf/cvpr/Chen0PLFR21} designs a non-blind network to restore night blurry images. Besides, Deng \etal \cite{DBLP:conf/iccv/DengRYWSC21} leverage a separable-patch architecture collaborating with a multi-scale integration scheme. For raw-based architecture, Xing \etal \cite{DBLP:conf/cvpr/XingE21} propose an end-to-end Joint Demosaicing and Denoising (JDD) network based on residual channel attention for joint demosaicing and denoising, and super-resolution. Wu \etal \cite{DBLP:journals/ijon/WuHDSZL21} advocate a blind JDD problem and invent a novel divide-and-conquer method to tackle with blind reconstruction from noisy raw images. Liu \etal \cite{DBLP:conf/cvpr/LiuQAJ0CG21} also devise Invertible Blocks to involve the learning of demosaicing. When concerning raw deblurring, Zhang \etal \cite{DBLP:conf/cvpr/ZhangDLK19} propose a stacked network consisting of several repetitive DMPHN modules to remove blurring from raw images. Liang \etal \cite{DBLP:journals/tmm/LiangCLH22} design a dual-branch network, which deals with color and spatial information in every blocks. Here, we refer to \cite{DBLP:conf/iccvw/YangSHC19,DBLP:journals/tmm/LiangCLH22} and utilize as the BFP module in the projection stage.

\subsection{Raw Image Pairs}
Raw restoration aims to recover high-quality raw images from the degraded raw image, while learning-based models need paired data. Hence, the acquirement of raw image pairs and efficient utilization of paired data are essential. It is time-consuming and labor-intensive to get them from the real world. Most algorithms \cite{DBLP:conf/cvpr/BaoYWBL20, DBLP:journals/corr/abs-1810-12575} add synthetic noises to clean images to generate pairs. Note that sRGB synthetic noise is not realistic enough and statistical noise like AWGN is too simple. Shot-and-read noise based on learning and other kinds of learned noise are more realistic instead. The others add shot-and-read noise to clean raw images and feed them into the ISP pipeline to learn the noise changing and extortion. CycleISP \cite{DBLP:conf/cvpr/ZamirAKHK0020} injects shot-and-read noise in raw images along with learned device-agnostic transformation rather than statistical synthetic noise. Tim \etal \cite{DBLP:conf/cvpr/BrooksMXCSB19} present a network to 'unprocess' images by reverting the process of ISP and add shot-and-read noise in the 'unprocessing' process to generate realistic raw sensor measurements. Linh \etal \cite{DBLP:conf/accv/LinhNA20} add shot-and-read noise to clean raw data and use GAN to generate realistic noisy raw images. Pseudo-ISP \cite{DBLP:journals/corr/abs-2103-10234} jointly learns ISP pipeline and signal-dependent 'rawRGB' noise model to synthesize realistic noisy images. Besides, paired raw images for deblurring can be acquired more easily. The blur-sharp image pairs are taken from the same scene using the sharper one as the ground-truth. In \cite{DBLP:journals/tmm/LiangCLH22}, the authors propose a raw deblurring dataset named Deblur-Raw. Many successive raw frames (3-5 frames) are randomly picked from raw videos in various scenes under the same camera setting, where the center one is used as the sharp ground-truth and the rest are blurred images. Considering that we already have some well-collected datasets, our goal is to reach good performances both in quality and quantity. Hence, our ELMformer focuses on how to efficiently and effectively utilize limited raw image pairs rather than extending data scale.

\begin{figure}[]
	\centering
	\includegraphics[width=0.9\linewidth] {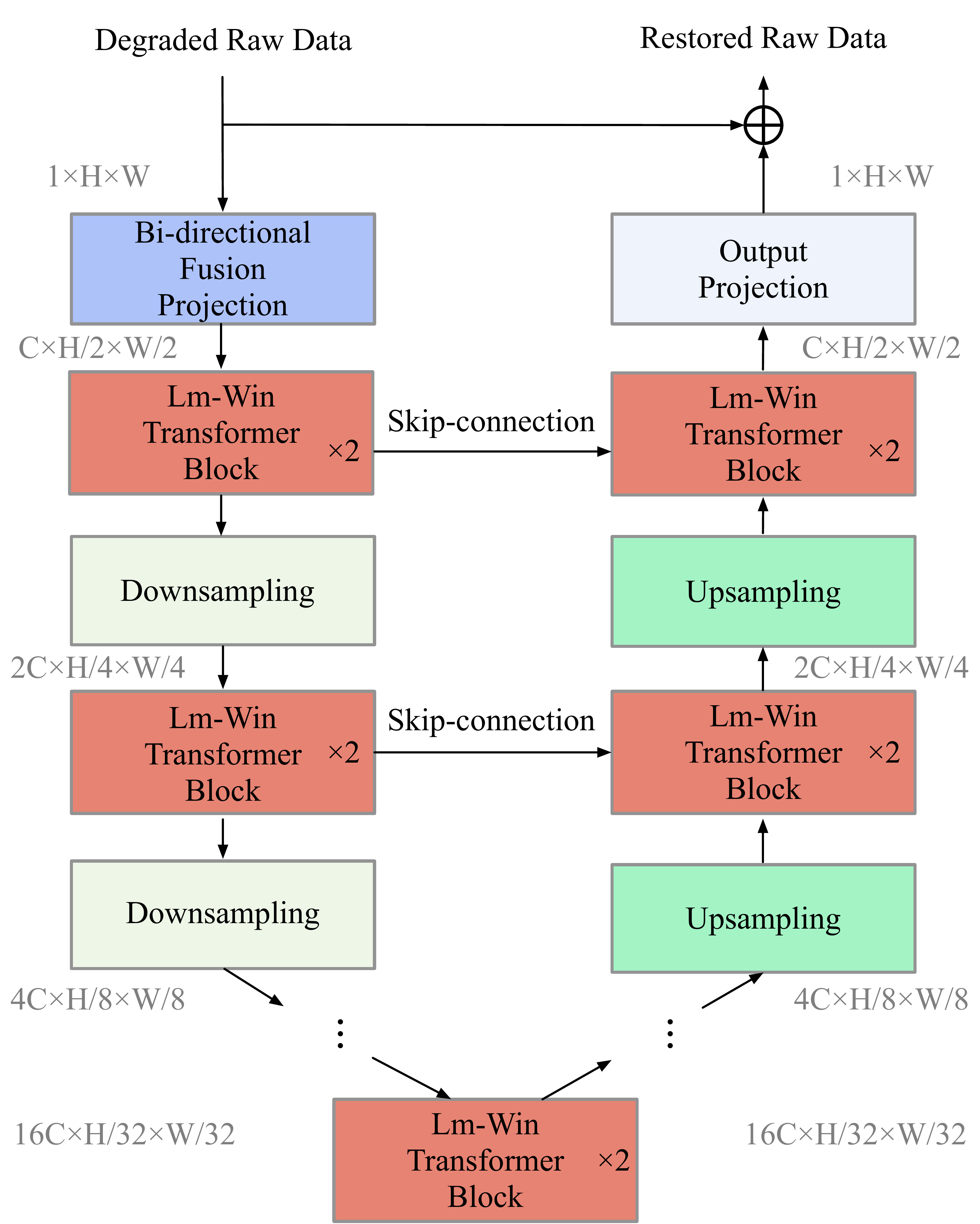}
	\caption{Illustration of the ELMformer pipeline.}
	\label{fig:pipeline-U}
\end{figure}

\subsection{Vision Transformer}
Transformer has been a powerful tool in computer vision recently. The global self-attention is found to be not only suitable for NLP but also CV tasks. For basic CV tasks, PVT \cite{DBLP:journals/corr/abs-2102-12122}, Swin Transformer \cite{DBLP:journals/corr/abs-2103-14030}, Vit\cite{DBLP:conf/iclr/DosovitskiyB0WZ21} and IPT \cite{DBLP:conf/cvpr/Chen000DLMX0021} attain SOTA performances. Specially, Uformer \cite{DBLP:journals/corr/abs-2106-03106} and SwinIR \cite{DBLP:journals/corr/abs-2108-10257} rise as cutting-edge methods in vision transformer for low-level vision tasks such as sRGB image denoising and deblurring. Uformer \cite{DBLP:journals/corr/abs-2106-03106} exploits Local-enhanced Window (LeWin) Transformer block to conduct local transformer operation on multi-scale feature maps which can simultaneously capture long-range features from high-level maps and low-level features from the local information provided by window shifting. Liang \etal \cite{DBLP:journals/corr/abs-2108-10257} organize several Swin Transformer layers together with a residual connection to form a Residual Swin Transformer Block to implement deep feature extraction for image restoration. Image synthesis and image editing are also relevant to our restoration tasks. Cao \etal \cite{DBLP:conf/nips/CaoHLWXFX21} propose image Local Autoregressive Transformer to better facilitate the locally guided image synthesis. Esser \etal \cite{DBLP:conf/cvpr/EsserRO21} combine convolution and Transformer to enhance the locality of high-resolution image synthesis. In this paper, our ELMformer exploits short-range dependencies from raw images, especially the L-MSA can minimize the performance drop with less computational cost.

\section{Method}

\subsection{Network Architecture}
Here, we apply the classic U-shaped architecture with the BFP module and Lm-Win Transformer blocks. As shown in Fig. \ref{fig:pipeline-U}, the overall structure of the proposed ELMformer is a hierarchical network with skip-connections between the encoder and the decoder. 

For instance, given a degraded raw data $I \in \mathbb{R}^{1 \times H \times W}$, ELMformer firstly applies BFP module to extract low-level features $X_0 \in \mathbb{R}^{C \times \frac{H}{2} \times \frac{W}{2}}$ and project single channel to diverse subspaces. Then features are feeded to $K$ encoders just like \cite{DBLP:conf/miccai/RonnebergerFB15,DBLP:conf/cvpr/IsolaZZE17}. Each encoder contains two sequential Lm-Win Transformer blocks and one downsampling layer. In the downsampling layer, we first reshape the flattened features into 2D spatial feature maps, and then downsample maps and double the channels using $4 \times 4$ convolution with a stride of 2. Then two Lm-Win Transformer blocks are added at the bottleneck stage. Due to the hierarchical structure, our Transformer blocks can capture longer and even global dependencies when the feature map size is reduced to the window size. The number of decoders is consistent with $K$ encoders. Each decoder also consists of one upsampling layer and two sequential Lm-Win Transformer blocks. $2 \times 2$ transposed convolution with a stride of 2 is used for upsampling. The skip connections are set in every stages. Finally, the features are flatten to 2D feature maps, and we upsample and project them to obtain a single-channel data $R \in \mathbb{R}^{1 \times H \times W}$. Note that we add a residual connection here as $I = I + R$, and it accelerates the learning speed of the whole network. $K$ is chosen as 4 to simply construct the whole pipeline.

\begin{figure}[]
	\centering
	\includegraphics[width=0.88\linewidth] {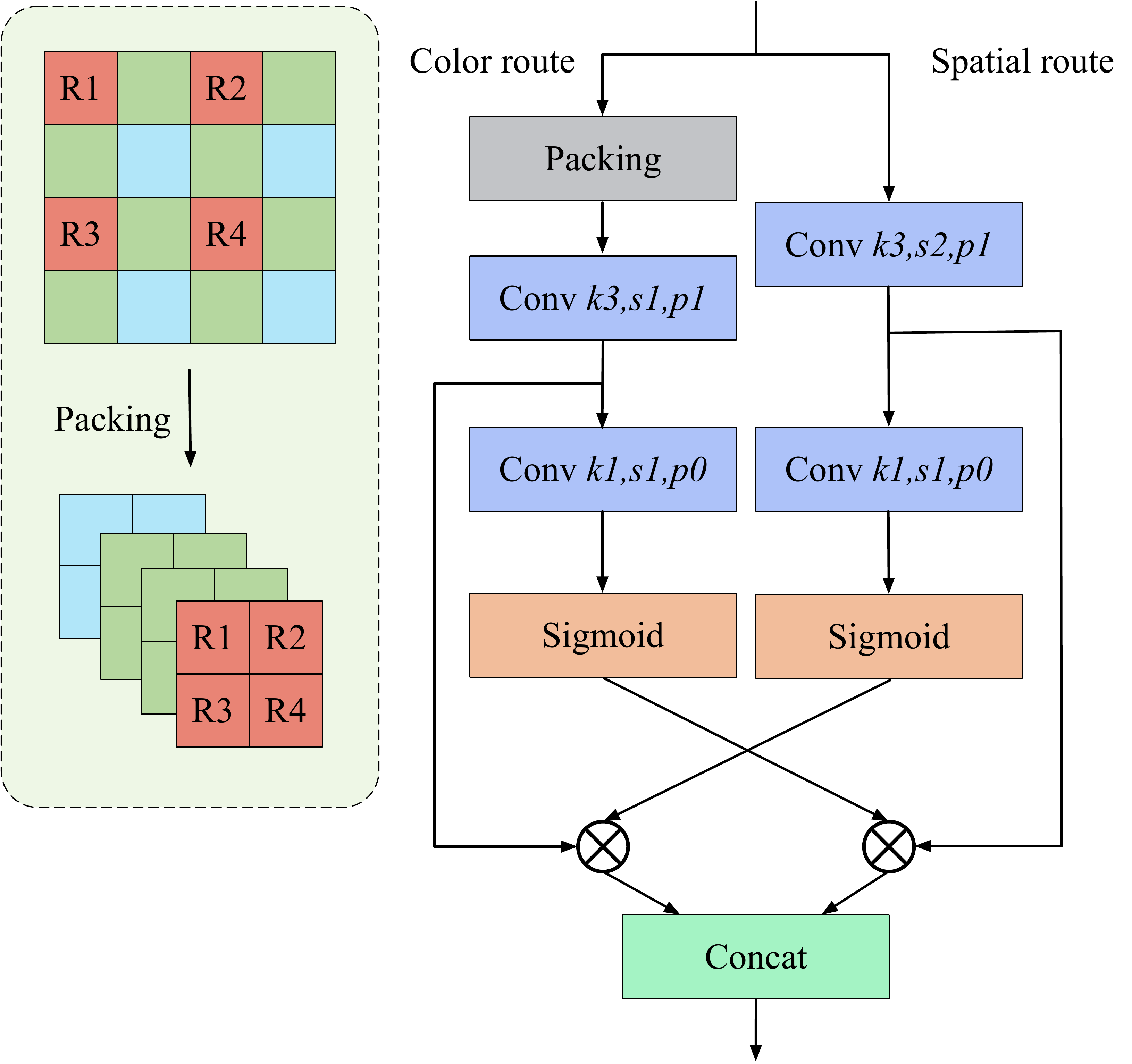}
	\caption{Illustration of the BFP module. $k, s, p$ represents kernel size, stride and padding, respectively.}
	\label{fig:BFP}
\end{figure}

\subsection{Bi-directional Fusion Projection}
Denote one raw image as $I  \in \mathbb{R}^{1 \times H \times W}$, though our goal is to restore a better raw image, we still need to demosaic it to get the final sRGB image. Hence, the correlations inside Bayer patterns which are essential should be preserved well. For most of existing works, they all split the one channel raw image into four channels (RGGB) according to the order of CFA. The strategy is called 'packing' and results in one tensor which is $4 \times \frac{H}{2} \times \frac{W}{2}$ with separate colors. Considering that raw images contain both color (single pixel) and spatial (between neighbour pixels) information, the rough packing strategy will downscale the resolution of images and breakdown the tight spatial order inside neighbor pixels. To avoid this defect and fully utilize the color and spatial information, we refer to \cite{DBLP:conf/iccvw/YangSHC19,DBLP:journals/tmm/LiangCLH22} and utilize it as the Bi-directional Fusion Projection module in the projection stage. 

As is shown in Fig. \ref{fig:BFP}, the module leads to two routes: Color route for keeping the color consistency and Spatial route for maintaining the spatial structure of original raw images. For the Color route, we perform packing on single-channel raw images and feed them into one $3 \times 3$ convolution to get their color-based features. Then the features are passed to one $1 \times 1$ convolution along with Sigmoid function. For the Spatial route, we remove the packing strategy and use one $3 \times 3$ convolution to downsample the input to the same scale as the output of packing. After that, we perform element-wise multiplications on both routes, which cross the color and spatial routes. Finally, the color-based and spatial-based features are concatenated in channel dimension for the following steps.

Notice that the original size of raw image is $1 \times H \times W$, we actually project it into high-dimensional subspaces and downsample it to $\frac{1}{2}$, which are features in size of $C \times \frac{H}{2} \times \frac{W}{2}$. Hence, we can take those projected features in smaller size for further efficient processing, and get lower computational costs. Unlike simple Input Projection module which only contains $3 \times 3$ convolutions for feature extraction, we actually compute the attention of those routes and allocate their weights to another. By this cross multiplication direction, two routes can be enhanced by each other, and the final features contain two parts: one is dominant by color information and the other is dominant by spatial structure. The BFP module can be formulated as follows:

\begin{equation}
	\begin{split}
		\text{F}^{spatial} &= \text{DS}(I) \otimes \text{Sigmoid}(\text{Conv}_{1 \times 1}(\text{Conv}_{3 \times 3}(P(I)))) \\
		\text{F}^{color} &= \text{Conv}_{3 \times 3}(\text{P}(I)) \otimes \text{Sigmoid}(\text{Conv}_{1 \times 1}(\text{DS}(I))) \\
		\text{F}^{BFP} &= \text{Concat}(\text{F}^{spatial}, \text{F}^{color}) 
	\end{split}
\end{equation}

Here, $\text{DS}(\cdot)$, $\text{P}(\cdot)$ and $\text{Concat}(\cdot,\cdot)$ are 2$\times$ downsampling, packing and concatenation on channel dimension, respectively. For input of one raw image $I$, the BFP module outputs $\text{F}^{BFP}$ as projected features.The BFP module utilizes two different properties of raw images: the spatial route considers pixels in the same color filter and the color route maintains the spatial structure of original images.

\subsection{Locally Multiplicative Window Transformer block}
\begin{figure}[]
	\centering
	\includegraphics[width=0.78\linewidth] {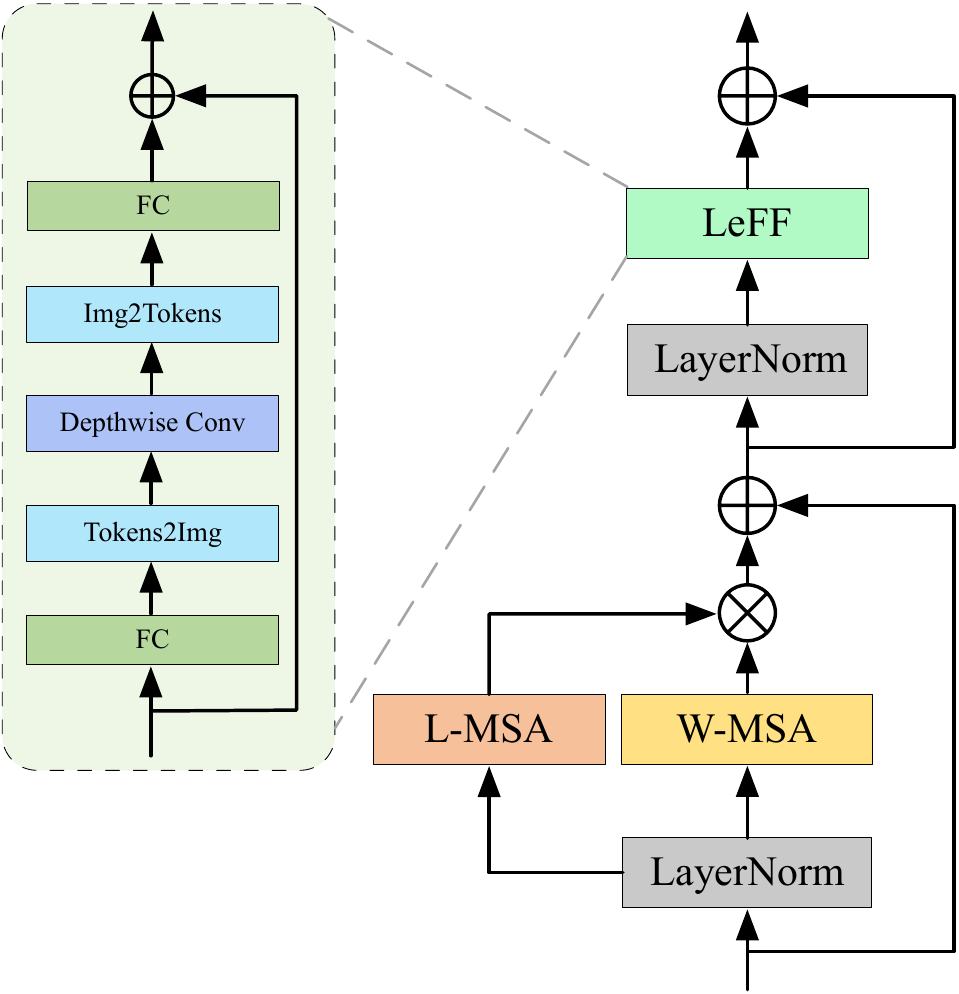}
	\caption{Details of the Lm-Win Transformer block.}
	\label{fig:Block}
\end{figure}

Standard vision transformer blocks \cite{DBLP:conf/iclr/DosovitskiyB0WZ21,DBLP:conf/nips/VaswaniSPUJGKP17} reshape images into vectors as tokens to compute global self-attention. To avoid the quadratic computation cost of self-attention on larger resolutions, Liu \etal \cite{DBLP:journals/corr/abs-2103-14030} propose a shifted window strategy. Although it brings the vision transformer into reality, many studies \cite{DBLP:journals/corr/abs-2103-15808,DBLP:journals/corr/abs-2104-05707} reveal that the local dependencies are not captured well by transformer. For low-level vision tasks such as denoising and deblurring, the local context information is essential due to the complementarity between a degraded pixel and its neighborhoods. Wang \etal \cite{DBLP:journals/corr/abs-2106-03106} enhance the ability of capturing locality by introducing the Locally-enhanced Feed-Forward Network (LeFF). It can replace the traditional MLP layer for better performance. Nevertheless, for raw data which is single-channel and high-resolution, the computational cost is still too large. What's more, though the packing strategy preserves the color information, the layouts of raw data implicit the spatial characteristics unlike images in sRGB space. Hence, it is essential to utilize the structural speciality of raw data in Transformer-based architectures.

To address the above mentioned issues, we propose a Locally multiplicative Window (Lm-Win) Transformer block, as shown in Fig. \ref{fig:Block}. The proposed Lm-Win Transformer block is effective and efficient by its special design for raw data, which not only captures long-range dependencies from the self-attention in transformer, but also involves the short-range connections into transformer to capture useful local context. The Self-Attention (SA) module reserves the original Window-based Multi-head Self-Attention (W-MSA) and extends one Locally Multiplicative Self-Attention (L-MSA) branch. For the Forward-Feedback Network (FFN) part, as many researches \cite{DBLP:journals/corr/abs-2104-05707,DBLP:journals/corr/abs-2106-13797} prove that the depth-wise $3 \times 3$ convolution can enhance the locality of transformer, we apply the same LeFF \cite{DBLP:journals/corr/abs-2106-03106}. Specifically, given the features at the $(l$-$1)$-th block $X_{l-1}$, the computation of a Lm-Win Transformer block is represented as:

\begin{equation}
	\begin{split}
		X^{SA}_l &= \textbf{W-MSA}(\textbf{LN}(X_{l-1}))\otimes \textbf{L-MSA}(\textbf{LN}(X_{l-1}))+X_{l-1} \\
		X_l &= \textbf{LeFF}(\textbf{LN}(X^{SA}_l))+X^{SA}_l \\
	\end{split}
\end{equation}

where $X^{SA}_l$ and $X_l$ are the outputs of the SA and LeFF parts respectively. $LN$ represents the layer normalization \cite{DBLP:journals/corr/BaKH16}.

\paragraph{\textbf{Revisiting Window-based Multi-head Self-Attention.}}
Unlike the vanilla Transformer which computes global self-attention, we perform self-attention within non-overlapping local windows \cite{DBLP:journals/corr/abs-2106-03106}. Given the 2D feature maps $X \in \mathbb{R}^{C \times H \times W}$ with $C$, $H$ and $W$ being the channel, height and width of the maps, we firstly split those feature maps into non-overlapping windows according to the $H$ and $W$ dimensions with the window size of $M \times M$. In this way, we get the flattened and transposed features $X_i \in \mathbb{R}^{M^2 \times C}$ from each window $i$. Suppose we perform self-attention on those flattened features in each window, and the head number is $k$ and the head dimension is $d_k = C/k$, the complexity of the $k$-th head self-attention in the non-overlapping windows should be: $O((M \times M)^2 \times d_k) = O(M^4 \times d_k)$.

Here, the $k$-th head self-attention in the non-overlapping windows can be defined as:
\begin{equation}
	\begin{split}
		X  &= \{X^1, X^2,...,X^N\}, N=\frac{H \times W}{M^2}\\
		Y^i_k &= \textbf{Attention}(X^iW^Q_k,X^iW^K_k,X^iW^V_k), i=1,2,...,N  \\
	\end{split}
\end{equation}
where $W^Q_k, W^K_k, W^V_k \in \mathbb{R}^{C \times {d_k}}$ represent the projection matrices of the queries, keys, and values for the $k$-th head, respectively.

Similar with \cite{DBLP:conf/naacl/ShawUV18,DBLP:journals/corr/abs-2103-14030}, we also choose the relative position encoding in the attention module, so its calculation can be formulated as:
\begin{equation}
	\textbf{Attention}(Q,K,V) = \textbf{Softmax}(\frac{QK^T}{\sqrt{d_k}}+B)V
\end{equation}

where $B$ is the relative position bias, whose values are taken from $\hat{B} \in \mathbb{R}^{(2M-1) \times (2M-1)}$ with learnable parameters.

\begin{figure}[]
	\centering
	\includegraphics[width=0.9\linewidth]{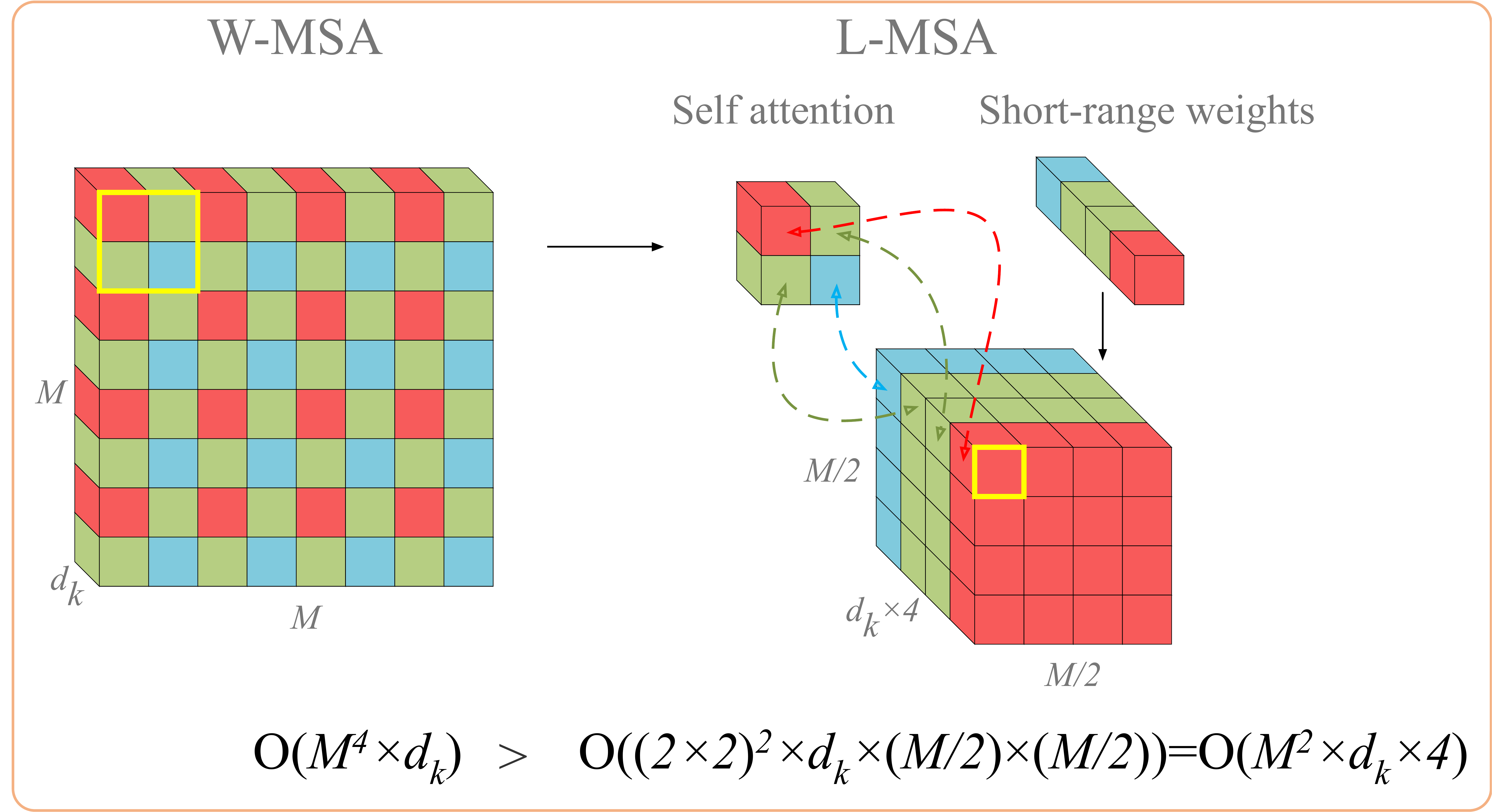}
	\caption{Illustration of costs between W-MSA and L-MSA.}
	\label{fig:binattn}
\end{figure}

\paragraph{\textbf{Locally Multiplicative Self-Attention.}}
Although W-MSA reduces the computational costs, the crude concatenation of nearby features is lack of consideration. For the local partial areas, the network should avoid learning dependencies from noisy pixels and give more attention to beneficial parts. So we propose the Locally Multiplicative Self-Attention (L-MSA) to enhance the color information and give a better explanation of self-attention.

As is shown in Fig. \ref{fig:binattn}, each pixel-wise location represents single color's feature. For ISP pipeline, neighborhood pixels are used for demosaicing to make up for the absence of the other color information of a pixel. Those adjacent homochromy pixels can also offer key contexts for those noisy or deficient points. Based on this insight, we select non-overlapping $2 \times 2$ sub-windows and rearrange each feature map according to sub-windows' relative positions.

This L-MSA is related to those $2 \times 2$ sub-windows. As is discussed previously, we can divide $\frac{M^2}{4}$ sub-windows for each window, so the total number of sub-windows is $\frac{HW}{4}$. For each sub-window in single window area, we compute its self-attention as what we illustrate before. 
 
\begin{equation}
	\begin{split}
		X  &= \{X_l^1, X_l^2,...,X_l^L\}, L=\frac{H \times W}{2^2}\\
		Z^j_k &= \textbf{Attention}(X_l^jP^Q_k,X_l^jP^K_k,X_l^jP^V_k), j=1,2,...,L  \\
	\end{split}
\end{equation}
where $P^Q_k, P^K_k, P^V_k \in \mathbb{R}^{C \times d_k}$ represent the projection matrices of the queries, keys, and values for the $k$-th head. Besides, $X^{\frac{M^2}{4}(i-1)+1}_l$, $X^{\frac{M^2}{4}(i-1)+2}_l$, ..., $X^{\frac{M^2}{4}i}_l$ are divided sub-windows from $X^i$, where $i = 1,2,...,N$.

Note that the computed self-attention reflects the weights of every color channel \ie Red, Green, Green and Blue (RGGB) in its corresponding position. Hence, we view them as short-range weights and multiply them with our W-MSA in every sub-window. For this part, we can represent it as:
\begin{equation}
	\begin{split}
		Z^i_k  &= \{Z^{\frac{M^2}{4}(i-1)+1}_k,Z^{\frac{M^2}{4}(i-1)+2}_k,...,Z^{\frac{M^2}{4}i}_k \}, i=1,2,...,N\\
		Y^i_k &= Y^i_k \cdot Z^i_k \\
		\hat{X_k}& = \{Y^1_k, Y^2_k,...,Y^N_k\}
	\end{split}
\end{equation}

$\hat{X}_k$ is the output of the $k$-th head. Then the outputs for all heads $\{1; 2; ... ; k\}$ are concatenated and then linearly projected to get the final result. In this way, we actually multiply W-MSA with L-MSA separately according to their CFA locations (RGGB).

By this Locally Multiplicative Self-Attention (L-MSA), all feature maps can be rearranged according to CFA-based structures, and focus more on the valid pixel-wise information. It makes up for the local connections in W-MSA, so our transformer block can tackle with severe noisy points. And the later experiments prove this conclusion. As is shown in Fig. \ref{fig:binattn}, the computation cost of L-MSA should be $ O(M^2 \times d_k \times 4)$. More detailed discussions about the computational cost of ELMformer are in Subsection \ref{computation}.

\begin{figure*}[]
	\centering
	\includegraphics[width=1\linewidth] {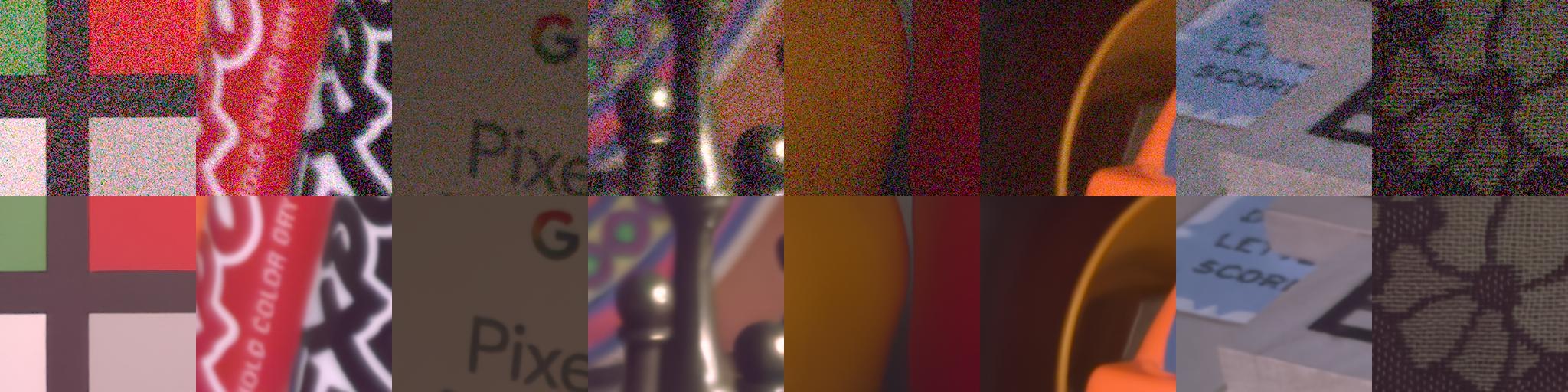}
	\caption{Real noise removal results of ELMformer on SIDD benchmark set. Raw images are fed into an official ISP pipeline for better visualization. Details can be seen when zoomed in. Up: noisy images; Down: denoised results.}
	\label{fig:sidd_bench}
\end{figure*}

\subsection{Computational Complexity}
\label{computation}
To give a clear comparison at the same scale, ELMformer and Uformer \cite{DBLP:journals/corr/abs-2106-03106} are designed with same numbers of blocks and projection features. Here, we theoretically analyze the computational complexity of the Lm-Win Transformer Block of ELMformer and the Le-Win Transformer Block of Uformer. For a batch size of single raw image $1 \times H \times W$, ELMformer firstly feeds it into the BFP module and gets one $C \times \frac{H}{2} \times \frac{W}{2}$ tensor, while Uformer results in one $C \times H \times W$ tensor. In the following Transformer blocks, our method only processes half-size tensors with the same channel number compared with Uformer. In regard to a single Transformer block under the same window size $M$, Lm-Win processes $\frac{HW}{4M^2}$ windows with each costing $O(M^4 \times d_k + M^2 \times d_k \times 4)$, and Le-Win processes $\frac{HW}{M^2}$ windows with each costing $ O(M^4 \times d_k)$. Hence, for a same raw image, Lm-Win costs $O(\frac{HWM^2 \times d_k}{4} + \frac{HW \times d_k}{M^2})$ while Le-Win needs $ O(HWM^2 \times d_k)$ for one raw image. Considering that both of them set $8$ as the window size, which is far smaller than normal size of unprocessed raw images ($thousands \times thousands$ in resolution), ELMformer costs nearly $4 \times$ less computational complexity compared with Uformer. As is shown in Fig. \ref{fig:top}, when we account into all factors of computation, ELFformer still remains at least one third FLOPs of Uformer, which means that ELMformer reaches higher results with less FLOPs. 

\subsection{Comparisons with Uformer}
Our ELMformer has a similar architecture to Uformer \cite{DBLP:journals/corr/abs-2106-03106} which is a U-net structure with vision transformer blocks. However, the Transformer blocks in ELMformer are different from that of Uformer. The Le-win Transformer block of Uformer consists of W-MSA module that conducts window-based self-attention and a forward-feeding network, while our Lm-Win Transformer block contains a novel L-MSA module designed especially for raw images. L-MSA conducts self-attention within subwindows in the non-overlapping shifted windows to produce self-attention weights of different color channels in RAW images. Then our Lm-Win Transformer block multiplies the output weights of L-MSA with the window-based self-attention outputs. The color channel self-attention weights of L-MSA serve as a reference for W-MSA and enhance the feature extraction ability of the Transformer blocks.

\section{Experiments}
\subsection{Datasets}
\textbf{Smartphone Image Denoising Dataset (SIDD):}
SIDD dataset \footnote{https://www.eecs.yorku.ca/~kamel/sidd/dataset.php} \cite{DBLP:conf/cvpr/AbdelhamedLB18} consists of 320 training image pairs (medium set) and 1280 image pairs (benchmark) for evaluations. These images are all collected with five specific smartphone cameras. The optical sensors of smartphones are of much smaller size compared with professional cameras, so the smartphone-produced images are more noisy but also have high resolution. 

\noindent\textbf{Darmstadt Noise Dataset (DND):}
DND dataset \footnote{https://noise.visinf.tu-darmstadt.de/downloads/} \cite{DBLP:conf/cvpr/PlotzR17} contains 50 noisy-clean image pairs from four consumer level cameras. Since the resolution of each image is very high, the dataset is cropped into patches of $512 \times 512$ and finally yields 1000 patches totally.

\noindent\textbf{Deblur-RAW Dataset}
Deblur-RAW Dataset \footnote{https://github.com/bob831009/raw\_image\_deblurring} \cite{DBLP:journals/tmm/LiangCLH22} contains 10252 RAW image pairs for raw deblurring, which contain blurred and corresponding deblurred ones. The whole dataset is split into 8752 and 1500 pairs, which are training and testing sets.

\begin{table}[]
	\centering
	\resizebox{!}{0.33\linewidth}
	{
		\begin{tabular}{c|c|c|c|c}
			\toprule
			\multirow{2}{*}{Method} &  \multicolumn{2}{c|}{PSNR} &
			\multicolumn{2}{c}{SSIM} \\ \cline{2-5}
			& r/r  & r/s  & r/r   & r/s	\\
			\midrule
			EPLL \cite{DBLP:conf/iccv/ZoranW11} & 40.73 & 25.19 & 0.935 & 0.842\\
			BM3D \cite{DBLP:journals/tip/DabovFKE07} &45.52&	30.95  &	0.980&	0.863  \\
			KSVD \cite{DBLP:journals/tsp/AharonEB06} & 43.26 & 27.41 & 0.969& 0.832\\
			WNNM \cite{DBLP:conf/cvpr/GuZZF14} &44.85	 & 29.54& 	0.975& 	0.888\\	
			FoE \cite{DBLP:journals/ijcv/RothB09} &43.13 & 27.18	 & 	0.969 & 0.812\\	
			TNRD \cite{DBLP:conf/cvpr/ChenYP15} & 42.77 &	26.99 &	0.945 & 0.744\\
			\midrule
			UPI* \cite{DBLP:conf/cvpr/BrooksMXCSB19}    & \textit{51.54}&  \textit{38.91}	& \textit{0.992}	& \textit{0.953}	\\
			UPI \cite{DBLP:conf/cvpr/BrooksMXCSB19}    &42.23&  28.39	& 0.888	& 	0.632		\\
			CycleISP* \cite{DBLP:conf/cvpr/ZamirAKHK0020}    & \textit{51.75} &  \textit{39.24}	&\textit{0.993}	& \textit{0.955}	\\
			CycleISP \cite{DBLP:conf/cvpr/ZamirAKHK0020}    & 47.98&  	35.02	& 	0.950	& 	0.846 		\\
			\midrule
			ELMformer   & \textbf{51.94} 	& \textbf{39.50} & \textbf{0.993}	&  \textbf{0.957}\\
			\bottomrule
		\end{tabular}
	}
	\caption{The average metrics on the SIDD benchmark set. *: results in \textit{italic} are from the official pretrained model trained by additional data.}
	\label{tab:SIDDbench}
\end{table}

\begin{table}[]
	\centering
	\resizebox{!}{0.33\linewidth}
	{
		\begin{tabular}{c|c|c|c|c}
			\toprule
			\multirow{2}{*}{Method} &  \multicolumn{2}{c|}{PSNR} &
			\multicolumn{2}{c}{SSIM} \\ \cline{2-5}
			& r/r  & r/s  & r/r   & r/s	\\
			\midrule
			EPLL \cite{DBLP:conf/iccv/ZoranW11} & 40.73 & 25.19 & 0.9350 & 0.8420\\
			BM3D \cite{DBLP:journals/tip/DabovFKE07}   &46.64&	37.78  & 0.9724&0.9308  \\
			KSVD \cite{DBLP:journals/tsp/AharonEB06} & 45.54 & 	36.59& 	0.9676 &  	0.9162 \\
			WNNM \cite{DBLP:conf/cvpr/GuZZF14} &46.30	 & 37.56	& 	0.9707& 	0.9313\\			
			FoE \cite{DBLP:journals/ijcv/RothB09} &45.78 & 35.99 & 0.9666 & 0.9042\\
			TNRD \cite{DBLP:conf/cvpr/ChenYP15} & 44.97 & 35.57 &0.9624& 0.8913\\
			\midrule 
			UPI* \cite{DBLP:conf/cvpr/BrooksMXCSB19}    & \textit{48.89} &  \textit{40.17}	& 	\textit{0.9824}	 	& \textit{0.9623}  	\\
			UPI \cite{DBLP:conf/cvpr/BrooksMXCSB19}    &48.44 & 39.47	&0.9802 &  0.9508	\\
			CycleISP* \cite{DBLP:conf/cvpr/ZamirAKHK0020}    & \textit{49.13} &  \textit{40.50} 	& \textit{0.9830}		& 		\textit{0.9655} 	\\
			CycleISP \cite{DBLP:conf/cvpr/ZamirAKHK0020}    & 48.75&  39.84	& 	0.9812	& 		0.9541	\\
			\midrule
			ELMformer    & \textbf{48.84}
            &  \textbf{40.06} & \textbf{0.9816}	&  	\textbf{0.9560}\\
			\bottomrule
		\end{tabular}
	}
	\caption{The average metrics on the DND dataset. *: results in \textit{italic} are from the official pretrained model trained by additional data.}
	\label{tab:DND}
\end{table}

\subsection{Experimental Settings}
Following the common training strategy of Transformer \cite{DBLP:conf/nips/VaswaniSPUJGKP17}, we use the AdamW optimizer \cite{DBLP:conf/iclr/LoshchilovH19} with momentum terms of (0:9; 0:999) and a weight decay of 0.02. We randomly augment the training samples using the horizontal flipping and rotating the images by 90 degrees, 180 degrees, or 270 degrees. We set the window size to $8 \times 8$ in Lm-Win Transformer blocks. For the convenience of downsampling, we stress that the window size in the bottleneck stage is set to $4 \times 4$. The projection dimension number is set to 32 for the BFP module. The cosine decay strategy is also applied to decrease the learning rate to $1e^{-6}$ with the initial learning rate $4e^{-4}$. The single $\ell_{1}$ loss is chosen as the loss function and we show the ablation study of loss functions in Appendix.

We apply PSNR and SSIM \cite{DBLP:journals/tip/WangBSS04} to evaluate the performance. These metrics are calculated in the Raw/Raw color space firstly. Then an ISP pipeline generates their corresponding sRGB images for evaluating in the Raw/sRGB color space. In the following sections and tables, 'r/r' and 'r/s' are short for 'Raw/Raw' and 'Raw/sRGB'. For SIDD and DND datasets, the whole evaluation processes are conducted online and only average PSNR and SSIM along with visualization of sample parts are the feedback. For Deblur-RAW dataset, we follow the default settings: conducting the training process on the training set and evaluating our model on the testing set.

\subsection{Compared Methods}
For a complete and fair comparison, we collect several denoising methods, which includes traditional blind ones, CNN-based ones and some simulated ISP pipelines. Besides, we collect several SOTA raw deblurring methods.

Of those denoising methods, EPLL \cite{DBLP:conf/iccv/ZoranW11}, BM3D \cite{DBLP:journals/tip/DabovFKE07}, KSVD \cite{DBLP:journals/tsp/AharonEB06}, WNNM \cite{DBLP:conf/cvpr/GuZZF14} and FoE \cite{DBLP:journals/ijcv/RothB09} are optimization based methods for denoising. TNRD \cite{DBLP:conf/cvpr/ChenYP15}, UPI \cite{DBLP:conf/cvpr/BrooksMXCSB19} and CycleISP \cite{DBLP:conf/cvpr/ZamirAKHK0020} are CNN-based methods. Besides, UPI \cite{DBLP:conf/cvpr/BrooksMXCSB19} and CycleISP \cite{DBLP:conf/cvpr/ZamirAKHK0020} also belong to ISP based pipelines which simulate the process of generating sRGB and raw images. Note that UPI \cite{DBLP:conf/cvpr/BrooksMXCSB19} and CycleISP \cite{DBLP:conf/cvpr/ZamirAKHK0020} need large-scale sRGB images to train the ISP pipeline, so we mark them with $*$ in Tab. \ref{tab:SIDDbench} and Tab. \ref{tab:DND}. For other methods without $*$, we follow the authors' instructions and train them only with SIDD medium set for a fair comparison.

Of those deblurring methods, DMCNN \cite{DBLP:conf/cvpr/NahKL17}, DeblurGAN \cite{DBLP:conf/cvpr/KupynBMMM18}, SRN \cite{DBLP:conf/cvpr/TaoGSWJ18}, UDS \cite{DBLP:conf/cvpr/LuCC19}, SDNet4 \cite{DBLP:conf/cvpr/ZhangDLK19}, DMPHN\_rgb \cite{DBLP:conf/cvpr/ZhangDLK19} are designed for sRGB images, so we only collect their resuls on sRGB color space. For DMPHN\_raw \cite{DBLP:conf/cvpr/ZhangDLK19} and RID \cite{DBLP:journals/tmm/LiangCLH22}, we evaluate them both on raw and sRGB color space.

\begin{table}[]
	\centering
	\resizebox{!}{0.3\linewidth}
	{
		\begin{tabular}{c|c|c|c|c}
			\toprule
			\multirow{2}{*}{Method} &  \multicolumn{2}{c|}{PSNR} &
			\multicolumn{2}{c}{SSIM} \\ \cline{2-5}
			& r/r  & r/s  & r/r   & r/s	\\
			\midrule
			DMCNN \cite{DBLP:conf/cvpr/NahKL17}   &-	   & 27.85  & -  & 0.880 \\
			DeblurGAN \cite{DBLP:conf/cvpr/KupynBMMM18}   &-   & 26.58  & - & 0.852 \\

			SRN \cite{DBLP:conf/cvpr/TaoGSWJ18}   &-   & 28.69  & -  & 0.925 \\

			UDS \cite{DBLP:conf/cvpr/LuCC19}   &-   & 24.60  & -  & 0.811 \\
			SDNet4 \cite{DBLP:conf/cvpr/ZhangDLK19}   &-   & 29.24 & -  & 0.920 \\
			DMPHN\_rgb \cite{DBLP:conf/cvpr/ZhangDLK19}   &-   & 28.73 & -  & 0.907 \\
			\midrule
			DMPHN\_raw \cite{DBLP:conf/cvpr/ZhangDLK19}   &41.68	   & 28.98  & 0.986  & 0.906 \\
			RID \cite{DBLP:journals/tmm/LiangCLH22}    & 42.71&  29.80		& 	0.989	& 	\textbf{0.929}		\\
			\midrule
			ELMformer     & \textbf{43.45} 	& \textbf{31.60} & \textbf{0.995}	&  0.913\\
			\bottomrule
		\end{tabular}
	}
	\caption{The average metrics on the Deblur-RAW testing set.}
	\label{tab:Deblur-RAW}
\end{table}

\subsection{Quantitative Results}
Here, we first compare our approach with the aforementioned SOTAs on both denoising benchmarks: SIDD and DND, and then evaluate on one deblurring benchmark: Deblur-RAW. Most results are from the official leaderboard on their websites or corresponding papers. Only some missing results are tested with authors' pretrained models and submitted to the online server for evaluation. For all three benchmarks, we record the PSNR and SSIM on Raw/Raw and Raw/sRGB space. Considering that only SIDD and Deblur-Raw have training sets, we evaluate DND by the model trained on SIDD medium dataset with no more external data. 

From Tab. \ref{tab:SIDDbench}, our ELMformer reaches the highest scores on all the metrics. ELMformer reaches 51.94 dB on PSNR and even surpasses $\text{UPI}^{*}$ \cite{DBLP:conf/cvpr/BrooksMXCSB19} and $\text{CycleISP}^{*}$ \cite{DBLP:conf/cvpr/ZamirAKHK0020} which are trained on huge additional data. The ELMformer has at least 3.96 dB and 4.48 dB improvements over CycleISP on raw and sRGB color space, respectively. Hence, our ELMformer can outperform other methods on raw denoising from smartphones. 

The results illustrated in Tab. \ref{tab:DND} differ. ELMformer is superior to most of methods except $\text{UPI}^{*}$ and $\text{CycleISP}^{*}$. Our ELMformer only has 0.10 dB and 0.22 dB promotion over CycleISP on raw and sRGB color space, respectively. Notice that $\text{UPI}^{*}$ and $\text{CycleISP}^{*}$ are trained on other generated large-scale data, while ELMformer only needs several hundred pairs of raw data, we suppose that those generated raw images can further extend the generalization ability. ELMformer still has a superiority over other methods at the same scale of training data. Besides, from Tab. \ref{tab:SIDDbench} and Tab. \ref{tab:DND}, we can find that ELMformer which is designed for raw restoration has an edge on raw color space compared with other methods and this advantage is weaken on sRGB color space.

We also evaluate on the raw deblurring task in Tab. \ref{tab:Deblur-RAW}. It is obvious that those SOTAs trained on the sRGB color space have a gap over methods designed for raw deblurring. Even DMPHN\_raw \cite{DBLP:conf/cvpr/ZhangDLK19} has minor superiority over DMPHN\_rgb \cite{DBLP:conf/cvpr/ZhangDLK19}. For three raw deblurring methods, ELMformer still exceeds others in all metrics except SSIM in sRGB color space but outperforms RID a lot in PSNR.

\subsection{Qualitative Results}
Then we show some representative results from three datasets for visual comparison. The raw images are fed into existing ISP pipelines and illustrated in sRGB space.

From Fig. \ref{fig:sidd_bench}, our method removes noisy points in SIDD and keeps the details successfully. It indicates that ELMformer can deal with real noises from mobile devices perfectly. Due to the space limit, we also illustrate the results of DND and Deblur-RAW datasets in Appendix. For DND, ELMformer can still effectively remove real noises and maintain the structure of locality even without training set in DND images' domain. For deblurring, our ELMformer has better visual results in deblurring. It further proves that methods trained on raw images supply color and spatial information and lead to clear fine-grained details.

\begin{table}[]
	\centering
	\resizebox{!}{0.145\linewidth}
	{
		\begin{tabular}{c|c|c|c|c|c|c|c}
			\toprule
			\multicolumn{3}{c|}{Modules} &  \multicolumn{2}{c|}{PSNR} &
			\multicolumn{2}{c|}{SSIM} &\multirow{2}{*}{GFLOPs}\\ 
			\cline{0-6}
			W-MSA & BFP & L-MSA & r/r  & r/s  & r/r  & r/s	& \\
			\midrule
			\checkmark  & & &	48.81	& 40.04	& \textbf{0.9817}	& 0.9557 & 10.24\\
			\checkmark  & \checkmark&	&	48.79	& 40.02  & 0.9816	&  0.9555& \textbf{2.84} \\
            \checkmark  & &\checkmark	&\textbf{48.85}	& \textbf{40.06}   & \textbf{0.9817} & 0.9556& 13.05\\
			\checkmark  & \checkmark&\checkmark& 48.84 	&  \textbf{40.06} & 0.9816	&  \textbf{0.9560} & 3.55\\
			\bottomrule
		\end{tabular}
	}
	\caption{Ablation study of modules on the DND dataset.}
	\label{tab:ablationDND}
\end{table}

\subsection{Ablation Studies}
To evaluate the effectiveness of individual parts in our ELMformer, we perform ablation study on denoising tasks and show the average PSNR and SSIM. As is shown in Table. \ref{tab:ablationDND}, we take W-MSA \cite{DBLP:journals/corr/abs-2106-03106} as basic blocks of our baseline and compare it along with BFP and L-MSA in the same U-shaped architecture. 

Comparing the first and second rows, it is confirmed that W-MSA can perform better with more computational cost as we expect. It reflects that although the FLOPs is nearly quartered, BFP module almost make up the drop in performance by fusion of color architecture and spatial information. Then when it comes to the first and third rows, we can find that L-MSA improves performance by enhancing short-range dependencies between local neighbor pixels while adding little computational cost. At last, the fourth row represents our proposed ELMformer. Compared with other three rows, ELMformer keeps a balance between performance and computational cost. It outperforms the baseline in all metrics while keeps only 1/3 FLOPs. By integrating BFP and L-MSA modules, ELMformer achieves the better results and loweer cost. More ablation studies are in Appendix.

\begin{figure}[t]
	\centering
	\includegraphics[width= 1\linewidth] {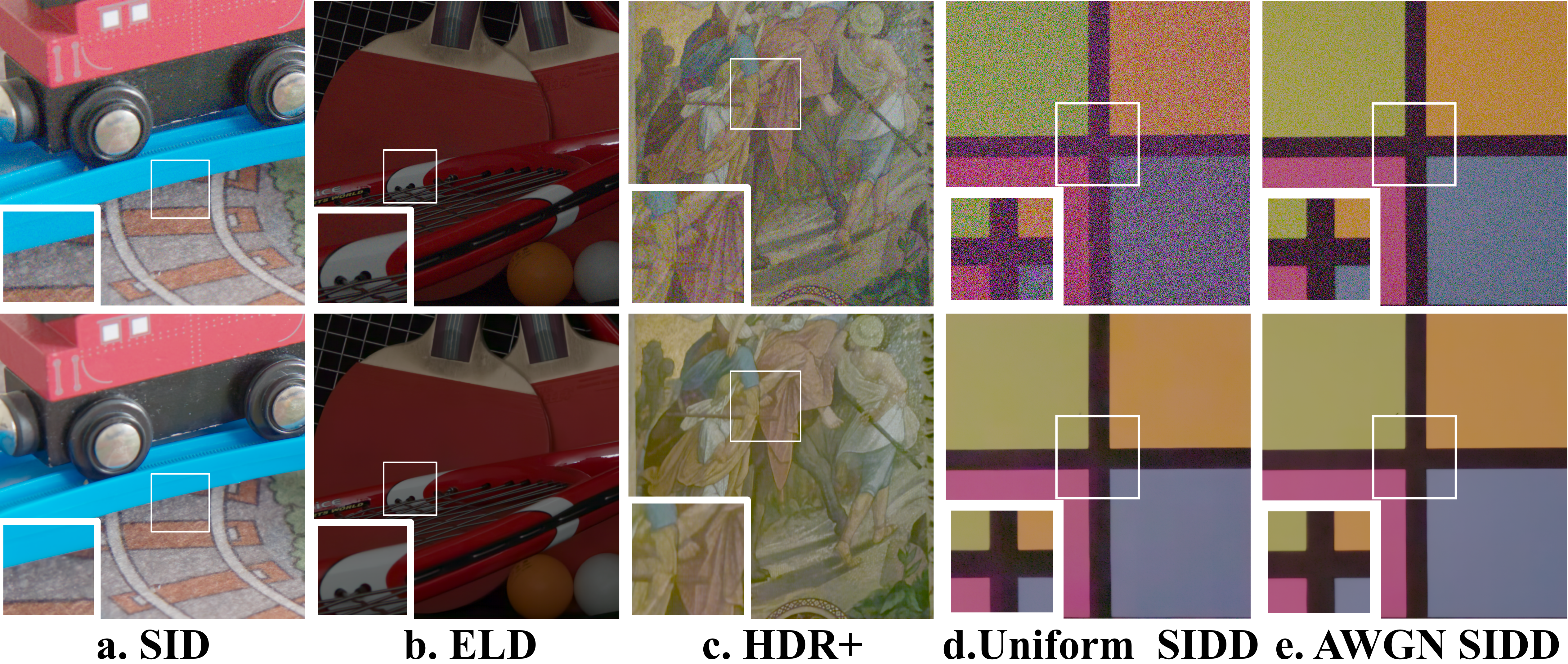}
	\caption{The denoising illustration of ELMformer on SID, ELD, HDR+, and two synthetic noisy SIDD datasets.}
	\label{fig:general}
\end{figure}

\subsection{Generalization Test}
Here, we firstly evaluate our denoising model on some extreme conditions to prove its powerful generalization ability. We select SID \cite{DBLP:conf/cvpr/ChenCXK18}, ELD \cite{DBLP:conf/cvpr/WeiFY020} and HDR+ \cite{DBLP:journals/tog/AdamsL16} datasets to separately show the denoising ability in short exposed, extremely dark and high-dynamic range conditions. Then we also add some synthetic noises to show its modeling ability. To be specific, we add AWGN and Uniform noises to the clean raw data of SIDD \cite{DBLP:conf/cvpr/AbdelhamedLB18} as the noisy ones, and directly apply our pretrained model of SIDD medium set to denoise. 

In Fig. \ref{fig:general}, we show the noisy and restored images which are processed by a simple ISP pipeline  \footnote{https://github.com/AbdoKamel/simple-camera-pipeline}. Compared with the noisy images, our method can produce noise-free outputs of high quality in several extreme conditions and maintain local details simultaneously. Also, we denoise some noisy raw images taken by mobile devices, \ie OPPO and XiaoMi, and the visualized results are in Appendix.

\section{Conclusions}
In this paper, we propose an efficient locally multiplicative Transformer called ELMformer for raw restoration. In contrast to existing CNN-based structures, our model build upon the BFP module and Lm-Win Transformer blocks. Two core components: BFP and L-MSA, are designed for bi-directional (color and spatial) feature fusion in the projection stage and efficient self-attention with stronger short-range connections. Extensive experiments demonstrate that ELMformer achieves competitive results on the raw denoising and deblurring tasks with less computation cost. More qualitative results from several datasets reveal that our pretrained model maintains a strong generalization ability.

\section*{ACKNOWLEDGMENTS}
This work was supported by the National Natural Science Foundation of China under Grant 62122060, the Special Fund of Hubei Luojia Laboratory under Grant 220100014, and the Fundamental Research Funds for the Central Universities under Grant 2042021kf0196. The numerical calculations in this work had been supported by the supercomputing system in the Supercomputing Center of Wuhan University.

\bibliographystyle{ACM-Reference-Format}
\balance 
\bibliography{sample-base}

\newpage
\appendix

\section{Ablation Studies}
Here, we further examine the effectiveness of the channel dimension $C$ in Tab. \ref{tab:ablationDim}. $C$ denotes the channel dimension of projection module. 

It is noteworthy that the computational cost will increase a lot with the increase of channel number, while the performance promotion is not significant. It reflects that 32 is a trade-off for the BFP module which can bring effective performance gains while maintaining low FLOPs.

\begin{table}[h]
	\centering
	\resizebox{!}{0.17\linewidth}
	{
		\begin{tabular}{c|c|c|c|c|c}
			\toprule
			\multirow{2}{*}{$C$} &  \multicolumn{2}{c|}{PSNR} &
			\multicolumn{2}{c|}{SSIM} & \multirow{2}{*}{GFLOPs} \\ \cline{2-5}
			& r/r  & r/s  & r/r   & r/s	&\\
			\midrule
			$16$ & 51.84	& 39.37	&  0.993	&  0.956&0.91\\
			$32$ & 51.94  	& 39.50 &  0.993 	&  0.957&3.55\\
			$48$ & 52.00 	& 39.56 &  0.993	&  0.957&7.91\\
			$64$ & 52.03 	& 39.60	&  0.993 	&  0.957&14.01\\
			\bottomrule
		\end{tabular}
	}
	\caption{Ablation studies of projection dimension on SIDD benchmark.}
	\label{tab:ablationDim}
\end{table}

To choose the optimal loss function for our ELMformer, we perform ablation study on SIDD benchmark. As is shown in Table. \ref{tab:ablationLoss}, we apply $\ell_{1}$, $\ell_{2}$ and $Charbonnier$ \cite{DBLP:conf/icip/CharbonnierBAB94} loss as alternative losses to supervise the training satge, and show the average PSNR and SSIM. Here, $Char$ is short for the $Charbonnier$ loss. 

It is clear that $\ell_{2}$ performs worst among those loss functions. It reflects that the errors in restoration tasks should not be enlarged. Comparing the first and third rows, $\ell_{1}$ loss shows a slight promotion than $Charbonnier$ loss in PSNR. Hence, we finally choose $\ell_{1}$ loss to supervise the training stage.

\begin{table}[h]
	\centering
	\resizebox{!}{0.155\linewidth}
	{
		\begin{tabular}{c|c|c|c|c}
			\toprule
			\multirow{2}{*}{$Losses$} &  \multicolumn{2}{c|}{PSNR} &
			\multicolumn{2}{c}{SSIM} \\ \cline{2-5}
			& r/r  & r/s  & r/r   & r/s	\\
			\midrule
			$\ell_{1}$ &	\textbf{51.94}	&\textbf{39.50}	&\textbf{0.993}&\textbf{0.957} \\
            $\ell_{2}$	&51.64 &39.05& \textbf{0.993}	& 0.954\\
            $Char$ & 51.93 	& 39.47 & \textbf{0.993}	&  \textbf{0.957}\\
           
			\bottomrule
		\end{tabular}
	}
	\caption{Ablation studies of losses on SIDD benchmark.}
	\label{tab:ablationLoss}
\end{table}

\section{Generalization Test in Extreme Conditions}
In this section, we evaluate our denoising model on some extreme conditions to prove its powerful generalization ability. We collect some noisy raw images taken by mobile devices, \ie Vivo and XiaoMi. Then we apply ELMformer to denoise those images and visualize them.

\begin{figure}[!b]
	\centering
	\includegraphics[width= 1\linewidth] {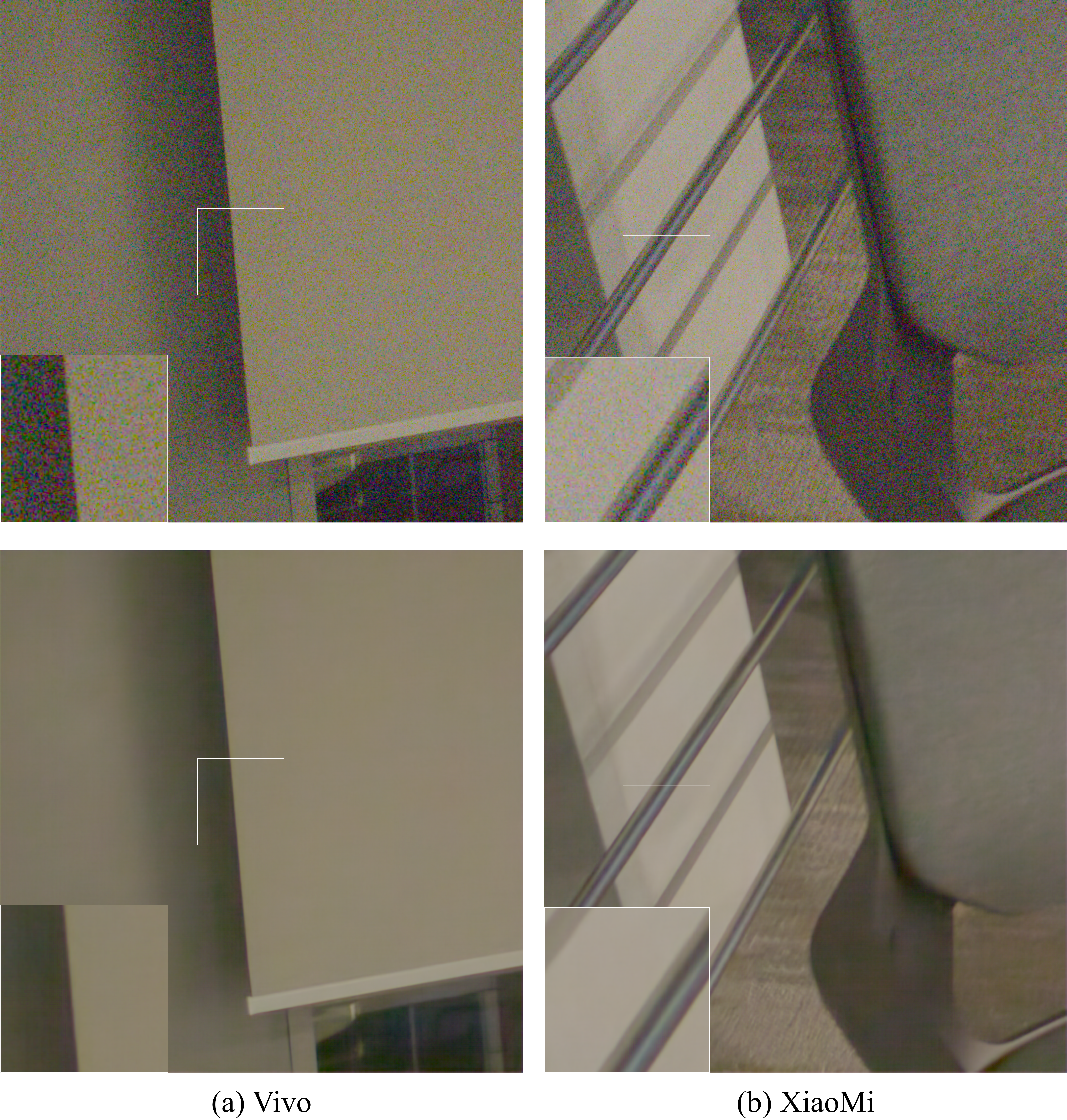}
	\caption{Real noise removal results of ELMformer on Vivo and XiaoMi. All raw images are fed into an existing ISP pipeline for better visualization.}
	\label{fig:mobile}
\end{figure}

Here, we also try our model on other raw images captured by mobile devices from other brands, such as Vivo and XiaoMi. Notice that both SIDD and DND do not contain data from Vivo and XiaoMi, and sensors result in different domains of raw images. From Fig. \ref{fig:mobile}, we observe that noisy points can be reduced well even without any training data from this domain.

\begin{figure*}[]
	\centering
	\includegraphics[width= 0.76\linewidth] {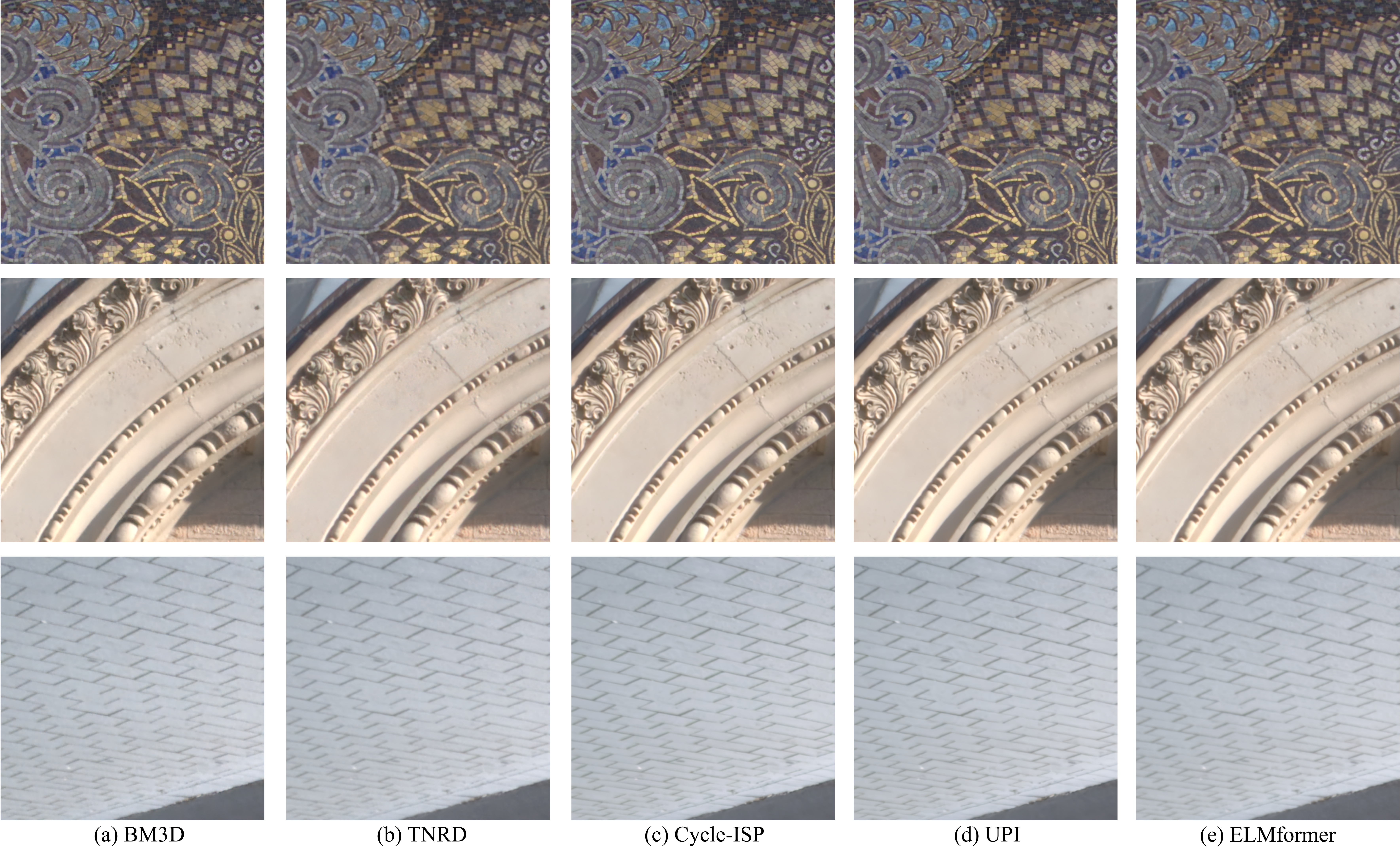}
	\caption{Real noise removal results on DND dataset. Raw images are fed into an existing ISP pipeline for better visualization.}
	\label{fig:dnd1}
\end{figure*}

\section{More visual results}

\begin{figure*}[]
	\centering
	\includegraphics[width= 0.78\linewidth] {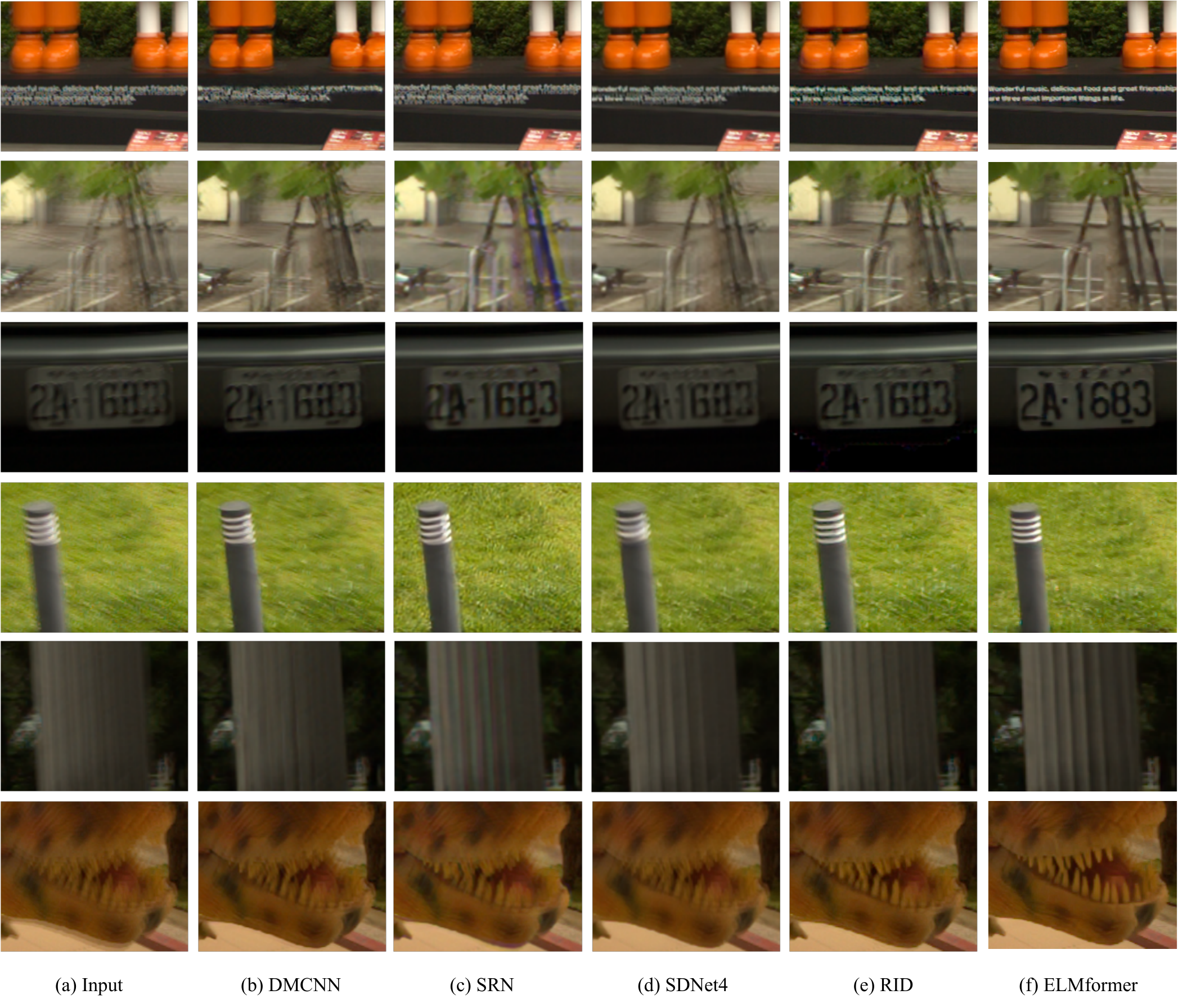}
	\caption{Deblur results on Deblur-Raw dataset. Raw images are fed into an existing ISP pipeline for better visualization.}
	\label{fig:deblur}
\end{figure*}

Here, we illustrate more denoising and deblurring visual results of our methods and SOTAs for comparison. In Fig. \ref{fig:dnd1}, those raw images are processed by official ISP pipelines online, and we collect their feedback in RGB images. For Fig. \ref{fig:deblur}, we also apply a simple ISP to raw images. To give an intuitive impression, we try our best to maintain the originality of images with fine-grained parts.

For denoising in Fig. \ref{fig:dnd1}, ELMformer can still effectively remove real noises and maintain the structure of locality even without training set in DND images' domain. It indicates that ELMformer can deal with real noises from mobile devices perfectly.

For deblurring in Fig. \ref{fig:deblur}, ELMformer has better visual results than other methods. It further proves that methods trained on raw images supply color and spatial information and lead to clear fine-grained details.

\end{document}